%% file: main.tex
\documentclass[10pt,twocolumn,letterpaper]{article}

\usepackage[]{cvpr}               
\input{preamble}

\newcommand{\TaskName}{\textit{\textbf{EchoFoley}}}{}

\newcommand{\BenchName}{\textit{\textcolor{olive}{\textbf{EchoFoley-6k}}}}{}

\newcommand{\ModelName}{\textcolor{violet}{$\mathit{EchoVidia}$}}{}

\definecolor{cvprblue}{rgb}{0.21,0.49,0.74}
\definecolor{mygray}{gray}{.9}
\usepackage[table]{xcolor}
\usepackage[pagebackref,breaklinks,colorlinks,allcolors=cvprblue]{hyperref}

\usepackage{algorithm}
\usepackage{algorithmic}
\usepackage{makecell}
\usepackage{multirow}
\usepackage{wrapfig}
\usepackage{booktabs}

\newcommand{\thickhline}{%
    \noalign {\ifnum 0=`}\fi \hrule height 1pt
    \futurelet \reserved@a \@xhline
}

\usepackage{tcolorbox}
\tcbset{
    colback=gray!3,
    colframe=black!40,
    arc=3mm,
    boxrule=0.3pt,
    left=4pt, right=4pt, top=4pt, bottom=4pt
}

\usepackage{titletoc}
\newcommand{\appendixtableofcontents}{
  \clearpage
  \onecolumn
  \begin{center}
    \centering
    \vspace*{4pt}
    \begin{tcolorbox}[
        width=0.92\textwidth,
        colback=white,
        colframe=black,
        boxrule=0.7pt,
        arc=2pt,
        left=10pt,
        right=10pt,
        top=8pt,
        bottom=8pt,
      ]
      \centering
      {\large\bfseries Appendix Contents}\\[2pt]
      \startcontents[appendix]
      \printcontents[appendix]{l}{1}{}
    \end{tcolorbox}
    \vspace{10pt}
  \end{center}
}

\def\paperID{36375} 
\def\confName{CVPR}
\def\confYear{2026}

\title{EchoFoley: Event-Centric Hierarchical Control for \\Video Grounded Creative Sound Generation}

\author{\textbf{Bingxuan Li$^{2,1,4}$, Yiming Cui$^{1}$, Yicheng He$^{1}$, Yiwei Wang$^{3}$, Shu Zhang$^{1}$, Longyin Wen$^{1}$, Yulei Niu\thanks{Project lead \& corresponding author}$^{~1}$}
\\
~$^1$ByteDance Intelligent Creation  $^2$University of Illinois Urbana-Champaign \\ $^3$University of California, Merced  ~$^4$University of California, Los Angeles 
\\
\textit{bl61@illinois.edu, yulei.niu@bytedance.com}
\\\\
\url{https://echofoley.github.io/}}

\begin{document}


\maketitle
\begin{abstract}
\input{sec/00-abstract}
\end{abstract}
\input{sec/01-intro}
\input{sec/02-related_works}
\input{sec/03-task}
\input{sec/04-benchmark}

\input{sec/05-experiment}

\input{sec/06-method}
\input{sec/07-conclusion}

\section*{Acknowledgments}
We thank Andong Deng, Zhenfang Chen, Dawei Du, Sijie Zhu for their insightful feedback and discussion on the work. This work is partially supported by the U.S. National Science Foundation (NSF) Grant CRII 2451683, an NVIDIA Academic Grants Program, a U.S. Bank Academic Research Award, the University of California, Merced, and a UC Merced Faculty Research Award. 
The views and conclusions are those of the authors and do not necessarily reflect the official policy or position of the U.S. Government.

{
    \small
    \bibliographystyle{ieeenat_fullname}
    \bibliography{main}
}

\input{sec/suppl}

\end{document}

%% file: preamble.tex
\usepackage{lineno}



%% file: sec/00-abstract.tex
Sound effects build an essential layer of multimodal storytelling, shaping the emotional atmosphere and the narrative semantics of videos. Despite recent advancement in video–text–to–audio (VT2A), the current formulation faces three key limitations: (1) an imbalance between visual and textual conditioning that leads to visual dominance; (2) the absence of a concrete definition for fine-grained controllable generation; (3) weak instruction understanding and following, as existing datasets rely on brief categorical tags. 
To address these limitations, we introduce \TaskName{} (\textbf{E}vent-\textbf{C}entric \textbf{H}ierarchical c\textbf{O}ntrol), a new task designed for video-grounded sound generation with both event-level local control and hierarchical semantic control. Our symbolic representation for sounding events specifies when, what, and how each sound is produced within a video or instruction, enabling fine-grained controls like sound generation, insertion, and editing. To support this task, we construct \BenchName{}, a large-scale, expert-curated benchmark containing over 6,000 video–instruction–annotation triplets and 42,000 fine-grained sounding event annotations.
Building upon this foundation, we propose \ModelName{}, a sounding-event-centric agentic generation framework with slow-fast thinking strategy. Experiments show that \ModelName{} surpasses recent VT2A models by 40.7\% in controllability and 12.5\% in perceptual quality.

%% file: sec/01-intro.tex
\section{Introduction}
\label{sec:introduction}

\begin{figure}
    \centering
    \includegraphics[width=1\linewidth]{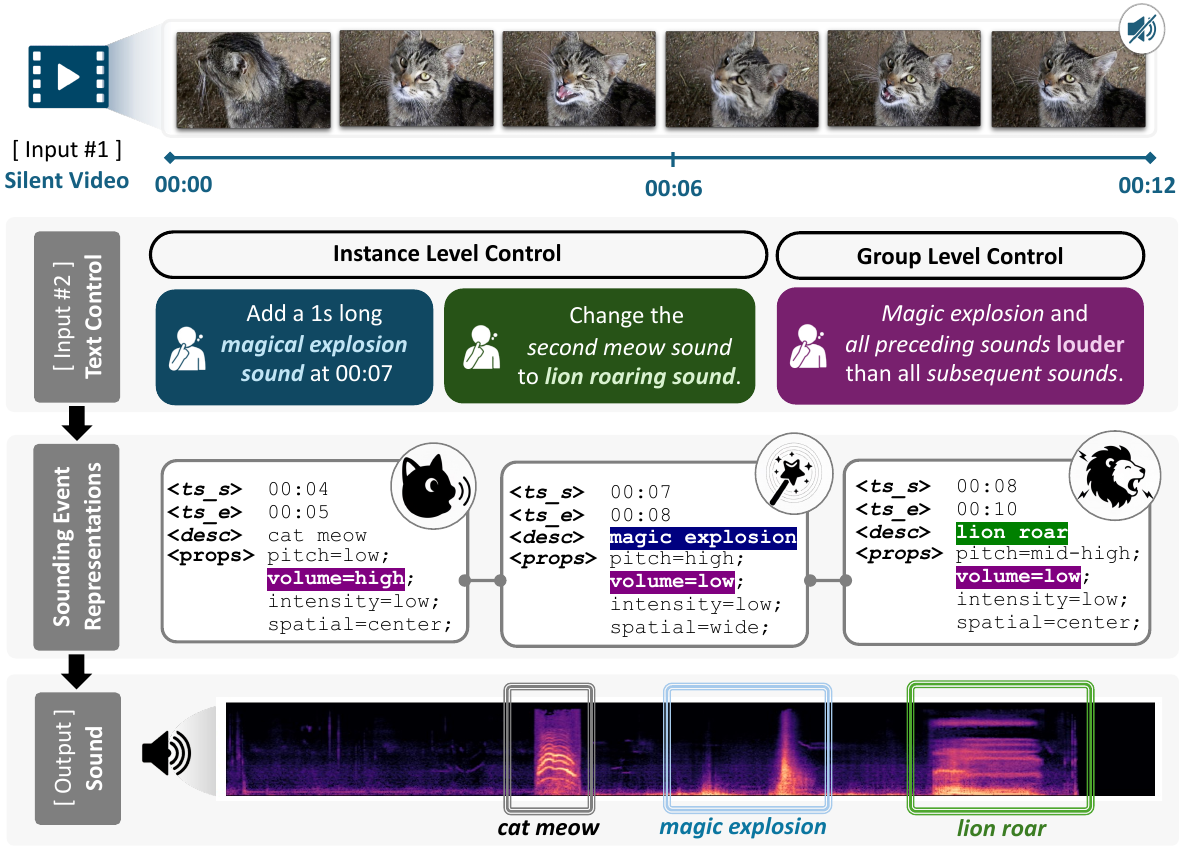}
    \vspace{-0.6cm}
    \caption{\textbf{Motivation of \textit{\TaskName}}. In creative storytelling, sound shapes the story we perceive. Given a silent video, generating audio that matches this story-shaped imagination requires fine-grained control over how each sound is crafted and transformed over time. We formulate such task with event-centric hierarchical control (\textit{e.g.}, instance control, group control), and propose effective solution.}
    \vspace{-0.8cm}
    \label{fig:teaser}
\end{figure}

Creative intelligence drives imagination to conceive stories and worlds beyond what we see. Recent advances in generative modeling have made this imagination increasingly tangible, allowing humans to use natural language to control highly realistic images and videos generation ~\citep{saharia2022imagen, sora, moviegen}, even enabling counterfactual world synthesis. Yet, the ultimate virtual worlds we create go beyond silent films, and sound represents a crucial dimension of creativity. 
Besides, controllability is essential for creative content synthesis, including video-to-audio generation. A slight change in sound can entirely recast a visual scene, reshaping how audiences perceive motion, story, and atmosphere. As the example illustrated in Figure~\ref{fig:teaser}, consider a silent video where a cat meows twice, by adding a magical whoosh and transforming the second meow into a lion’s roar, this ordinary scene tells a creative story: ``\textit{A cat greeted me with a gentle meow, then with my wizard magic, the cat can now suddenly roar with the mighty voice of lion.}'' Generating such sound effect 
requires fine-grained controls beyond the video level.



While recent models have made progress toward controllable video-to-audio generation, existing approaches still fall short of the fine-grained creative control envisioned above. Current systems typically rely on optional text instructions 
to guide sound generation, such as short labels~\citep{zhang2024foleycrafter, chen2024multifoley, cheng2025mmaudio} (\textit{e.g.}, ``cat meowing'') or short natural language descriptions or reasoning ~\citep{liu2025thinksound, shan2025hunyuanvideo, rong2025audiogenie, xie2024sonicvisionlm, rowles2025foley} (\textit{e.g.}, ``distant rumble of thunder becomes louder and more intense''). However, the current coarse instruction setting cannot support fine-grained controls or editing, especially for complex visual and temporal scenes. First, these video-level instructions operate at the granularity of a specific sound category and thus cannot distinguish multiple events of the same category. For example, ``change the cat's meow to a louder sound'' is ambiguous when the video contains several meow events. The instruction cannot specify which meow—\textit{e.g.}, ``the second meow at 00:07''—should be replaced with a lion roar or edited independently of the others. Second, the instructions on sound effects are limited to single aspects, for example the timbre of the sound (\textit{e.g.}, ``add a magic explosion''). Such instructions cannot support editing multiple attributes, like a combination of timbres, orders, durations, and volumes, at the same time (\textit{e.g.}, ``insert a 1-second magic explosion at 00:07 and make this explosion and all preceding sounds louder than all sub-sequent ones''). These limitations hindered the development of fine-grained controllable video-text-to-audio generation.

In this work, we present \TaskName{}, a new task designed for \textbf{E}vent-\textbf{C}entric and \textbf{H}ierarchical c\textbf{O}ntrol in video-grounded sound generation. Instead of video-level control, we focus on event-level control to disentangle a target sounding event from others. We further introduce a symbolic sounding event representation to define structure the event, enabling a hierarchical control to edit sounding effects according to a sound category, a single sounding event, or its specific attributes.
To support this task, we construct \BenchName{}, a high-quality, expertly curated benchmark comprising over 6,000 video–instruction–audio triplets with 42,000 densely annotated sounding events. Each example combines natural-language instructions about how sounds should evolve with event-level temporal annotations indicating when and where each event occurs.


While many VT2A methods have achieved impressive results, our preliminary studies on \TaskName{} reveal that they remain limited in how they represent and control auditory imagination through natural language. To bridge this gap, we further propose \ModelName{}, a training-free agentic framework  with slow-fast thinking.
\ModelName{} enables hierarchical and interpretable auditory control—allowing models to reason about events, understand human intent, and generate creative, contextually aligned soundscapes. Experiments show that \ModelName{} outperforms recent VT2A baselines by 40.7\% in controllability and 12.5\% in perceptual quality, demonstrating its effectiveness in fine-grained, instruction-guided sound generation.


In summary, the main contributions of our work are:
\begin{enumerate}
\item We formulate \TaskName{}, introducing a new paradigm for event-centric hierarchical control in video-grounded sound generation, and define a symbolic sounding-event representation that explicitly specifies when, what, and how each sound is produced within a video.
\item We construct \BenchName{}, a large-scale, densely-annotated, and expertly-curated benchmark containing over 6,000 video–instruction-annotation triplets of data and over 42,000 dense sounding event annotations. We also provide a suite of metrics to support systematic evaluation of the proposed task.
\item We propose \ModelName{}, a sounding-event-centric agentic generation framework with slow-fast thinking that enables fine-grained control. Experiments demonstrate significant improvements in controllability, semantic alignment, and quality over recent VT2A baselines.
\end{enumerate}

%% file: sec/02-related_works.tex
\section{Related Work}
\label{sec:related_works}

\noindent{\textbf{Audio-Video Correspondent Datasets.}} 
Learning robust multimodal representations relies on large-scale datasets that align visual and auditory signals. 
\textit{VGGSound}~\citep{vggsound} first established broad correspondences between videos and sound events, while \textit{ego4Dsounds}~\citep{ego4Dsounds}, \textit{AVVP}~\citep{tian2020avvp}, and \textit{ASVA}~\citep{linz2024asva} extended this paradigm to egocentric, weakly supervised, and synchronized settings for richer temporal and spatial modeling. Audio–text corpora enable grounding of auditory semantics in language. \textit{AudioSet}~\citep{audioset} and the \textit{BBC Sound Effects Library}~\citep{BBCSoundEffects} provide broad sound category coverage, while \textit{AudioCaps}~\citep{audiocaps} and \textit{WavCaps}~\citep{wavcaps} pair audio clips with natural language captions for multimodal alignment. Our work advances the direction with fine-grained event-level correlations.

\noindent{\textbf{Multimodal Conditioned Audio Generation.}} 
Recent works demonstrate how cross-modal reasoning enables controllable multimodal content generation~\citep{li2025metal, zhou2025contrastive, zhao2024reffly}. 
In the multimodality-conditioned audio generation domain, diffusion- and transformer-based models\cite{zhang2025long, zhang2026foleycrafter, xu2024video}, including \textit{Seeing\&Hearing}~\citep{xing2024seeing}, \textit{Diff-Foley}~\citep{luo2023difffoley}, \textit{MultiFoley}~\citep{chen2024multifoley}, \textit{MMAudio}~\citep{cheng2025mmaudio}, \textit{HunyuanVideo-Foley}~\citep{shan2025hunyuanvideo}, \textit{Hear-Your-Click}~\citep{liang2025hearyourclick}, \textit{YingSound}~\citep{chen2024yingsound}, and \textit{ThinkSound}~\citep{liu2025thinksound} synthesize temporally aligned, video-conditioned audio. 
Yet current models remain over-optimized for visual alignment, struggle with complex multi-object scenes, and lack fine-grained controllability through text. Some recent works have attempted to adopt MLLMs for multi-stage V2A \cite{rowles2025foley, xie2024sonicvisionlm}. Our work evaluate and advance this direction to fine-grained control with symbolic representation.

\noindent{\textbf{Sounding Event Localization.}} 
A key step toward controllable audio generation is accurately localizing where and when sound events occur. 
Early work by \citet{tian2018audio} established the audio–visual event localization task. Subsequent studies advanced granularity and data scale: Hebbar \textit{et al.}~\citep{hebbar2023dataset} curated a movie-based detection dataset, while Mahmud and Marculescu~\citep{mahmud2023ave} introduced \textit{AVE-CLIP} with pre-trained audio–language representations for temporal segmentation. 
Recent works~\citep{gao2023collecting, fan2025fine, geng2025longvale, zhou2025towards} explore weakly supervised, long-video, and open-vocabulary settings for more flexible event reasoning. 
Egocentric settings further enrich perception by coupling actions and sounds in first-person videos~\citep{huang2023egocentric, chen2024soundingactions}. 
Our work extends this direction toward fine-grained controlled audio generation.

\noindent{\textbf{Video Event Reasoning with Large Multimodal Models.}}
Large multimodal models extend language understanding to spatiotemporal reasoning in videos. 
Models such as \textit{Video-LLaMA}~\citep{videollama}, \textit{Video-LLaVA}~\citep{lin2024video}, \textit{MovieChat}~\citep{moviechat}, \textit{VideoChat}~\citep{videochat}, \textit{Vidi}~\citep{vidi}, \textit{Vita}~\citep{vita}, and \textit{Gemini 2.5}~\citep{gemini} enable detailed video understanding, editing, and instruction following through unified visual–language representations. 
Earlier work such as \textit{Vid2Seq}~\citep{vid2seq} explored dense video captioning and temporal grounding, laying the foundation for reasoning-aware video–language alignment. 
Classical temporal reasoning and localization studies~\citep{Zhou_2018_ECCV, charades, localizing_moments, qvhighlight} further established the importance of identifying moments and actions from natural language queries. 
Together, these advances inspire the integration of multimodal reasoning and temporal grounding in our setting.

%% file: sec/03-task.tex
\section{Task: \textit{EchoFoley}}
\label{sec:benchmark}

\TaskName{} aims to provide video-grounded sound generation with fine-grained controls at event level. To support such controllability, we first introduce a symbolic representation for sounding events that serves as an intermediate interface between natural language instructions and video-grounded audio generation.
Based on this symbolic representation, we formulate an event-centric, hierarchically controllable video-grounded sound generation task.

\subsection{Sounding Event Representation}
\label{sec:sounding_event_representation}

We define \emph{sounding events} as temporally-localized audio segments corresponding to actions or objects that are grounded in either the video content and the instruction context. We formulate the symbolic representation of a sounding event $e$ as a structured tuple:
\[
e = (\mathbf{t},\, d,\, \mathbf{p}),
\]
where $\mathbf{t}=(t_{start},t_{end})$ denotes the temporal location of the event along the video timeline, 
$d$ is a semantic description about $<$subject, action, object$>$ where object is optional,
and 
$\mathbf{p}$ specifies controllable audio properties such as timbre, pitch, intensity, and spatialization.

\subsection{Task Definition}
\label{sec:task}

Building on the symbolic formulation above, we define \TaskName{} as producing audio tracks that reflect both the video context, and \emph{faithfully satisfies the event-centric hierarchical control constraints} specified by the user’s instruction. 
Given a video $V$ and an instruction $I$, the corresponding set of sounding events $\mathcal{C}$ is denoted as:
\[
\mathcal{C}=\{(\mathbf{t},\, d,\, \mathbf{p})|V,I\}
\]
The task reduced to video-audio generation if $I$ is \texttt{Null}. The instruction can explicitly or implicitly specify the temporal location (\textit{e.g.}, ``\texttt{1 second-long} magical explosion at time stamp \texttt{0:03}'', ``\texttt{second} meow sound''), semantic description (\textit{e.g.}, ``\texttt{cat meows}'', ``\texttt{ball hits bottles}''), and attributes (\textit{e.g.}, ``\texttt{low pitch}'', ``\texttt{soft timbre}''), and any-level of combination.

We further organize the symbolic control space into hierarchical levels of sounding-event–centric control, where each level governs events at a different level of semantic and temporal abstraction. This hierarchy comprises three \textit{control levels}:
\begin{itemize}
    \item \textbf{Instance Level} — controls the properties of a single sounding event, such as emission or insertion of an event. (\textit{e.g.}, ``change the \emph{second} meow into a lion roar'').
    \item \textbf{Group Level} — coordinates multiple related events, enabling control over interactions, co-occurring actions, or repeated event sequences. (\textit{e.g.}, ``transform \emph{all} cat meows in the video into lion-like vocalizations'').
    \item \textbf{Video Level} — shapes the overall acoustic profile, balance, and distribution of all events throughout the video. (\textit{e.g.}, ``render the whole soundtrack with a cartoon-like audio aesthetic'').
\end{itemize}

We also design three complementary \textit{control types} independent to the control levels:
\begin{itemize}
\item \textbf{Temporal Control} — determines \emph{when} a sounding event occurs and \emph{how long} it lasts, regulating timestamps and durations (\textit{e.g.}, ``delay the explosion by one second'').
\item \textbf{Timbre Control} — specifies \emph{what} an object should sound like by modifying auditory texture or identity (\textit{e.g.}, ``make the cat bark and the dog meow'').
\item \textbf{Volume Control} — adjusts \emph{how strong or distant} a sound appears, manipulating volume depth (\textit{e.g.}, ``make the thunder louder'').
\end{itemize}

This hierarchical design enables flexible and interpretable modulation of audio generation, from precise event-level adjustments to global scene-level control.

%% file: sec/04-benchmark.tex
\section{Benchmark: \textit{EchoFoley-6k}}

To enable systematic studies of the proposed task, we establish a comprehensive benchmark comprising (1) a large-scale, densely annotated dataset (\BenchName{}), and (2) an evaluation suite with both automatic and human assessments. In this section, we will introduce the dataset construction process (§\ref{sec:data_construction}), summarize its key statistics (§\ref{sec:data_stats}), and outline the evaluation protocols (§\ref{sec:evaluation_metrics}).

\subsection{Data Curation}
\label{sec:data_construction}

\begin{figure}[tb]
    \centering
    \includegraphics[width=\linewidth]{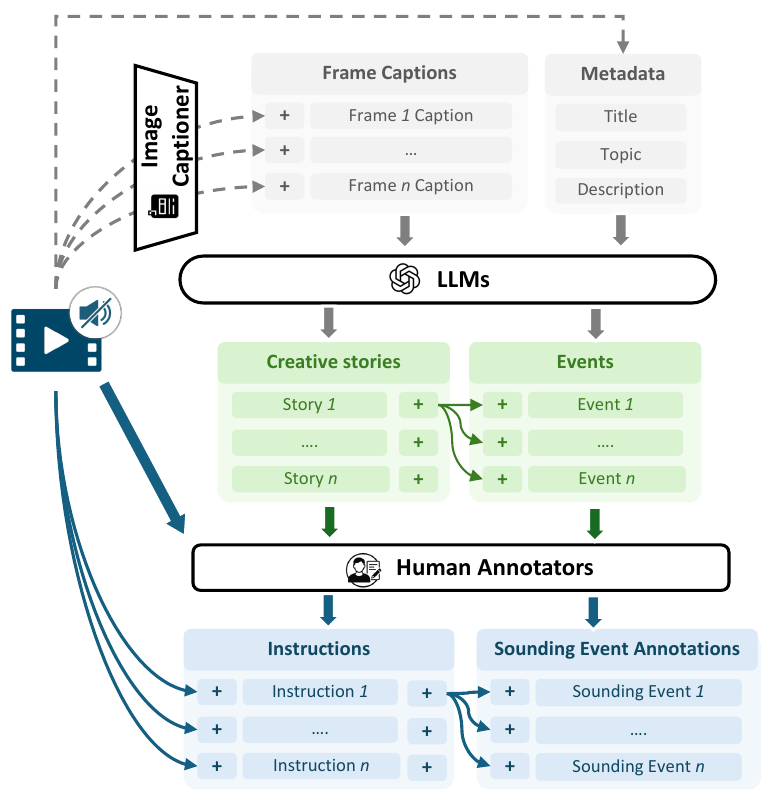}
    \caption{\textbf{Data Curation Pipeline }of \BenchName{}}
    \label{fig:data_pipeline}
    \vspace{-0.4cm}
\end{figure}

Figure~\ref{fig:data_pipeline} illustrates the data construction pipeline of \BenchName{}. We sample candidate videos from VGGSound~\citep{vggsound} and PE Video Dataset~\citep{PEVideo}, followed by a series of processing and annotation steps (please refer to appendix for implementation details):

\noindent\textbf{Step 1: Video Filtering.} We begin with motion-centered videos where sound-producing interactions are visually evident—for example, an animal vocalizing, an object being struck, or an environment changing. This ensures that sounding events are grounded in the visible scene rather than arbitrary background audio.

\noindent\textbf{Step 2: Metadata and Frame Captioning.} For each video, we collect coarse metadata (title and short description) and generate frame-level visual captions that describe how the scene evolves moment by moment. These captions provide structured visual grounding and temporal cues for both narration and event identification.

\noindent\textbf{Step 3: Story Proposal and Event Extractions.} Using the metadata and captions, a large language model produces an imaginative story describing how sound could shape the scene and proposes an initial set of sounding events (\textit{e.g.}, ``where a meow starts, or when an object impact occurs''). These serve as high-level scaffolds rather than final labels.

\noindent\textbf{Step 4: Human Modification.} Human annotators then convert the creative story into concrete, fine-grained \emph{instructions} describing how sounds should change over time (\textit{e.g.}, “make the second meow sharper and more excited”), and refine the list of candidate sounding events by adjusting temporal boundaries and specifying auditory attributes such as intensity, pitch, and texture. 

This process yields a dataset where each video is associated with multiple natural-language instructions and detailed sounding-event annotations.

\begin{table}
\centering
\small
\begin{tabular}{l c}
\toprule
\textbf{Statistic} & \textbf{Number} \\
\midrule
Total Number of Videos & 937 \\
Video Topics & 14 \\
Avg Video Duration & 11 \\
Video Duration Range & 6$\sim$30 seconds \\
\midrule
Total Sounding Events & 5622 \\
Avg. Sounding Events per Video & 6 \\
\midrule
Total Instructions & 6,018 \\
Avg. Instructions per Video & 12 \\
\bottomrule
\end{tabular}
\vspace{-0.2cm}
\caption{\textbf{Dataset Statistics for \BenchName{}.}}
\label{tab:dataset_stats}
\vspace{-0.2cm}
\end{table}

\subsection{Dataset Statistics}
\label{sec:data_stats}

\begin{figure}
    \centering
    \includegraphics[width=\linewidth]{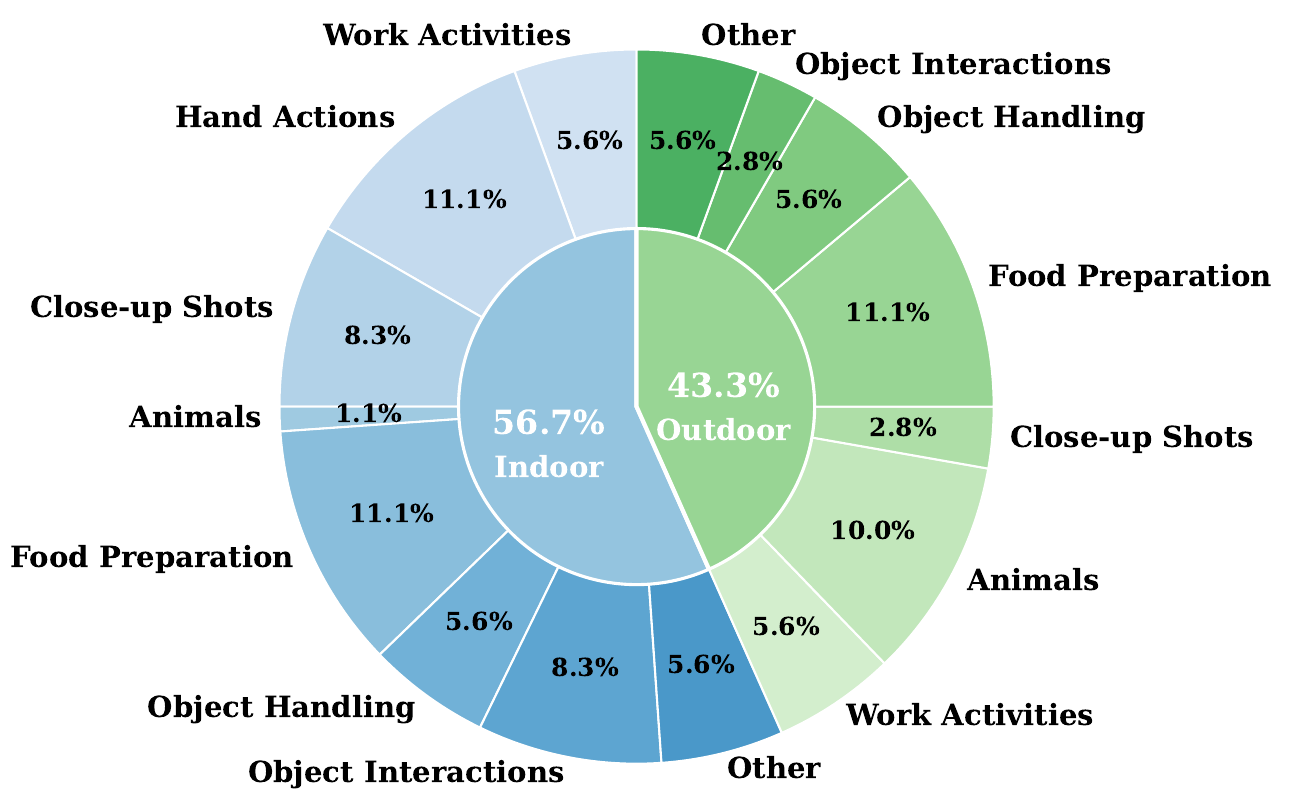}
    \vspace{-0.2cm}
    \caption{\textbf{Video Topic Distribution} in \BenchName{}}
    \label{fig:topic_distribution}
    \vspace{-0.6cm}
\end{figure}

Table~\ref{tab:dataset_stats} and Figure~\ref{fig:topic_distribution} summarize the overall scale and diversity of \BenchName{}. The benchmark comprises 6,000 video–instruction pairs and 42,000 fine-grained event annotations. Each video contains multiple instructions that provide temporal and attribute-level control cues, enabling detailed mapping between user intent and event-level sound synthesis.
Across the dataset, videos span a wide range of visual scenes and auditory contexts (Figure~\ref{fig:topic_distribution}). 
Table~\ref{tab:dataset_stats} summarizes the scale and statistics of \BenchName{}. 

\subsection{Evaluation Suite}
\label{sec:evaluation_metrics}

To assess objective performance, we design an evaluation suite with automatic and human evaluation metrics. More implementation details are in the appendix.

We design three automatic evaluation metrics to evaluate the controllability the event-level perspective, focusing when, what, and how the generated sound behaves.

\noindent\textbf{When: Temporal Control.}
For each event $e_i$, we compare the ground-truth event duration
$\mathbf{t}=(t_{start}, t_{end})$
with the predicted active region of the generated audio $\hat{\mathbf{t}}$
obtained from an AudioLLM-based onset/offset detector given the instruction.
To measure how well the predicted event location aligns with the ground truth, we compute the intersection-over-union (IoU) of the two intervals:
\[
\text{TempIoU}(e)
= \frac{\left| \mathbf{t} \cap \hat{\mathbf{t}} \right|}{\left| \mathbf{t} \cup \hat{\mathbf{t}} \right|}.
\]
The temporal controllability score for a single video is the average IoU across all events:
\[
\text{TempCtl}(V,I)
= \sum_{e\in\mathcal{C}} \text{TempIoU}(e),
\]
where $\mathcal{C}=\{e|V,I\}$ is the ground-truth sounding event set given the video $V$ and instruction $I$.

\noindent\textbf{What: Timbre Control.}
To assess whether the generated sound matches the intended semantic identity of event $e$, we compute the semantic similarity between the audio segment and the event’s semantic description $d$. The audio segment of the event is extracted from the video's original audio $A$, grounded by its temporal location $\textbf{t}$. We then calculate the CLAP similarity score~\cite{laionclap2023} between $A_\textbf{t}$ and $d$:
\[
\text{CLAP}(e) = \text{sim}\big(A_\textbf{t},\, d\big).
\]
The timbre controllability score for a single video is obtained by averaging CLAP scores across all the events:
\[
\text{TimbCtl}(V,I)
= \sum_{e\in\mathcal{C}} \text{CLAP}(e).
\]

\noindent\textbf{How: Volume Control.} Volume characterizes the perceptual attributes of a sound, reflecting contextual cues like strength, spatial distance, emotion, which should be aligned with video or instruction context. To eliminate the impact of reasonable volume variation, we define three loundness levels: [\textit{low}, \textit{medium}, \textit{high}]. To evaluate volume controllability, we compute the relative loudness of each generated audio segment $A_\textbf{t}$ to the entire audio sequence $A$ as $\ell_\textbf{t}$ by
\[
\ell_\textbf{t} =
\begin{cases}
\text{low}, &
\displaystyle \frac{\text{Loudness}(A_\textbf{t})}{\text{Loudness}(A)} < \tau_1,
\\[8pt]
\text{medium}, &
\displaystyle \tau_1 \le 
\frac{\text{Loudness}(A_\textbf{t})}{\text{Loudness}(A)} < \tau_2,
\\[8pt]
\text{high}, &
\displaystyle 
\frac{\text{Loudness}(A_\textbf{t})}{\text{Loudness}(A)} \ge \tau_2.
\end{cases}
\]

\noindent where \(\tau_1\) and \(\tau_2\) are predefined thresholds and \(\text{Loudness}(\cdot)\) is a function that extracts the volume of an audio segment.

We define the depth controllability score as the average equivalence between the loudness $\ell^{\text{gt}}_\textbf{t}$ of the ground-truth audio segment and $\ell_\textbf{t}$:
\[
\text{VolCtl}
= \sum_{e\in\mathcal{C}}\mathbf{1}\!\left[\, \ell_\textbf{t} = \ell^{\text{gt}}_\textbf{t} \,\right].
\]

To complement automatic metrics, we further conduct human evaluations to assess perceptual aspects that are difficult to capture algorithmically. Human raters score each generated audio clip on a five-point Likert scale along the following dimensions:

\begin{itemize}
    \item \textbf{Instruction Adherence:} How well the audio follows the user instruction, including requested changes in timing, timbre, or loudness.
    
    \item \textbf{Audio--Visual Coherence:} How consistent and synchronized the generated sounds are with the visual content, including object actions, motion, and event boundaries.

    \item \textbf{Perceptual Quality:} The overall naturalness, clarity, and realism of the audio as perceived in the video context.
\end{itemize}

%% file: sec/05-experiment.tex
\section{Evaluation}
\label{sec:experiment}

\begin{table*}[t]
\centering
\small
\begin{tabular}{lccccccccccc}
\toprule
\multirow{2}{*}{\textbf{Model}} &
\multicolumn{3}{c}{Automatic (Controllability)} &
\multicolumn{4}{c}{Automatic (Audio Quality)} &
\multicolumn{3}{c}{Human Evaluation} \\
\cmidrule(lr){2-4} \cmidrule(lr){5-8} \cmidrule(lr){9-11}
& \textbf{TempCtl}
& \textbf{TimbCtl}
& \textbf{VolCtl}
& \textbf{PQ}
& \textbf{PC}
& \textbf{CE}
& \textbf{CU}
& \textbf{Instr.}
& \textbf{A--V Coh.}
& \textbf{Qual.} \\
\midrule

MMAudio-S-16kHz

& 0.26 & 0.21 & 0.52
& 6.22 & 2.79 & 3.18 & 5.69
& 1.60 & 3.20 & 3.07 \\

MMAudio-S-44.1kHz

& 0.30 & 0.24 & 0.55
& 6.25 & 3.00 & 3.21 & 5.47
& 2.00 & 3.53 & 3.13 \\

MMAudio-M-44.1kHz

& 0.28 & 0.23 & 0.54
& 6.06 & 2.79 & 3.11 & 4.88
& 1.60 & 3.60 & 3.13 \\

MMAudio-L-44.1kHz

& 0.29 & 0.23 & 0.56
& 5.97 & 2.84 & 2.99 & 5.08
& 1.93 & 3.53 & 3.13 \\

ThinkSound

& 0.18 & 0.34 & 0.50
& 6.49 & 3.03 & 3.45 & \underline{5.94}
& 1.53 & 2.20 & 2.00 \\

AudioGenie

& 0.27 & 0.23 & 0.58
& 6.22 & 2.79 & 3.18 & 5.69
& 1.47 & 3.47 & 3.47 \\

HunyuanVideo-Foley-xl

& 0.41 & 0.46 & 0.67
& 6.47 & \underline{3.48} & 3.46 & 5.64
& \underline{2.60} & \textbf{4.20} & \underline{3.73} \\

HunyuanVideo-Foley-xxl

& \underline{0.43} & \underline{0.48} & \underline{0.69}
& \underline{6.49} & 3.45 & \underline{3.49} & 5.66
& 2.53 & 4.07 & 3.67 \\

\midrule

\textbf{\ModelName{}}
& \textbf{0.72} & \textbf{0.78} & \textbf{0.75}
& \textbf{7.32} & \textbf{4.29} & \textbf{4.33} & \textbf{6.50}
& \textbf{3.80} & \underline{3.93} & \textbf{3.79} \\

\bottomrule
\end{tabular}
\vspace{-0.2cm}
\caption{\textbf{Main Evaluation Results} with metrics for
Controllability (\textbf{TempCtl}: Temporal Controllability,
\textbf{TimbCtl}: Timbre Controllability,
\textbf{VolCtl}: Volume Controllability), Audio quality (\textbf{PQ}: Production Quality,
\textbf{PC}: Production Complexity,
\textbf{CE}: Content Enjoyment,
\textbf{CU}: Content Usefulness),
and human evaluation (\textbf{Instr.}: Instruction Adherence,
\textbf{A--V Coh.}: Audio--Visual Coherence,
\textbf{Qual.}: Perceptual Quality).
The best-performing value per metric is \textbf{bolded}, and the second-best value is \underline{underlined}.}
\label{tab:combined_results}
\vspace{-0.4cm}
\end{table*}

We evaluate current video-text-to-audio models on our \BenchName{} benchmark to test their (1) event-level hierarchical controllability and (2) generated audio quality via automatic and human assessment.

\subsection{Main Evaluation Setup}
\label{sec:experiment_setup}

\noindent\textbf{Models.}
We evaluated 8 recent open-source video-text-to-audio  generation models that support both visual and textual conditioning, including MMAudio~\citep{cheng2025mmaudio}, ThinkSound~\citep{liu2025thinksound},
AudioGenie~\citep{rong2025audiogenie},
and HunyuanVideo-Foley~\citep{shan2025hunyuanvideo}.

\noindent\textbf{Evaluation Metrics.}
We evaluate models with the \textit{proposed} automatic and human metrics introduced in Section~\ref{sec:evaluation_metrics}. For objective controllability, we report three automatic metrics: \textbf{TemporalCtl}, \textbf{TimbreCtl}, and \textbf{VolCtl}, which quantify accuracy in temporal alignment, timbre manipulation, and loudness modulation, respectively. For human evaluation, we use three subjective metrics: \textbf{Instruction Adherence}, \textbf{Audio--Visual Coherence}, and \textbf{Perceptual Quality}, which collectively measure controllability, cross-modal alignment, and perceptual naturalness of the generated audio.
To further assess the intrinsic quality of the generated audio, we adopt the \textbf{Audio Aesthetics Score (AES)}~\cite{aes}. AES decomposes audio aesthetics into four interpretable dimensions:
\begin{itemize}
    \item \textbf{Production Quality (PQ):} Technical fidelity, including clarity, dynamic range, freq. balance, and spatial realism.
    \item \textbf{Production Complexity (PC):} Richness and structural complexity of elements within the audio scene.
    \item \textbf{Content Enjoyment (CE):} Subjective enjoyability, reflecting artistic expressiveness, emotional impact, and listener engagement.
    \item \textbf{Content Usefulness (CU):} Practical utility of the audio, capturing its potential reusability in creative workflows.
\end{itemize}
For human evaluation, we randomly sample 50 video–instruction pairs and recruit 6 participants to rate each generated audio. The inter-annotator agreement for human evaluation is 0.62 (Cohen’s kappa).

\begin{figure}[tb]
    \centering
    \includegraphics[width=0.9\linewidth]{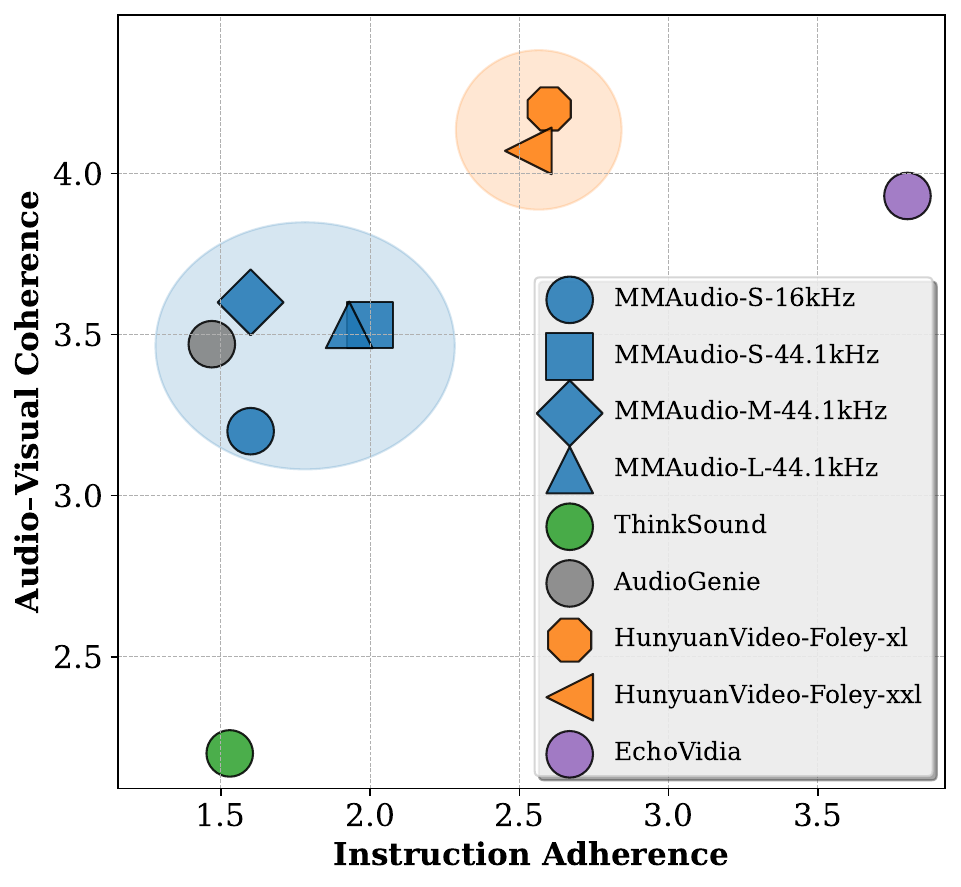}
    \vspace{-0.4cm}
    \caption{The relationship of Instruction Adherence ($x$-axis) plotted against Audio–Visual Coherence ($y$-axis) of models shows Visual Dominance Bias.}
    \label{fig:visualbias}
    \vspace{-0.5cm}
\end{figure}

\begin{figure}[tb]
    \centering
    \includegraphics[width=1\linewidth]{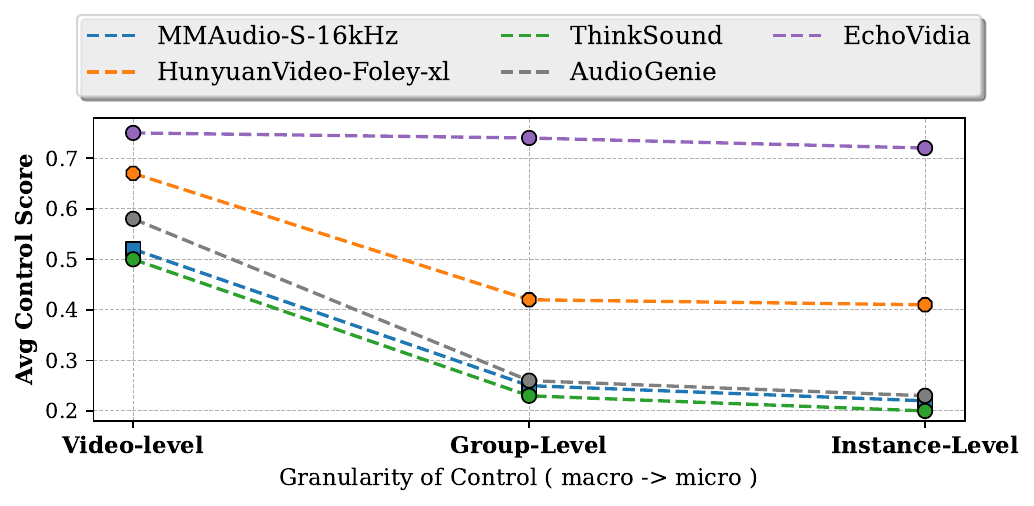}
    \vspace{-0.6cm}
    \caption{Average controllability scores for different levels of control granularity, from macro (global) to micro (atomic).}
    \label{fig:level}
    \vspace{-0.5cm}
\end{figure}

\subsection{Main Evaluation Results and Analysis}
\label{sec:experiment_results}

Table~\ref{tab:combined_results} summarizes the performance of existing VT2A  models across controllability, audio quality, and human evaluation metrics. We have three main insights as follows.

\noindent\textbf{Lack of Controllability.}
Overall, current VT2A models demonstrate significant limitations in controllability, particularly for temporal and timbre dimensions. Temporal controllability remains low across all the baseline models (0.18$\sim$0.43), revealing persistent difficulty in aligning generated audio with visual event timing. Timbre controllability shows similarly modest performance, with most models clustered around 0.21$\sim$0.24, and only ThinkSound (0.34) and HunyuanVideo-Foley variants (0.46$\sim$0.48) showing moderate gains. Volume controllability is comparatively higher (0.50$\sim$0.69), suggesting that loudness modulation is inherently easier for current systems, though still far from reliable. Human evaluations reveal similar weaknesses, where annotators consistently report low \emph{Instruction Adherence} (below 2.60 out of 5.0).


\noindent\textbf{Visual Dominance with Text Neglect.} Figure~\ref{fig:visualbias} reveals a consistent pattern across existing VT2A models: high Audio–Visual Coherence but low Instruction Adherence.
This imbalance indicates a strong visual dominance effect---VT2A models tend to rely heavily on visual cues while largely ignoring the fine-grained controls specified in the text instruction. Most models cluster toward the upper–left region of the plot, showing that they are able to synchronize generated audio with visible events reasonably well (\textit{e.g.}, Audio–Visual Coherence ranges in 3.2$\sim$3.6), yet fail to follow detailed textual directives (\textit{e.g.}, Instruction Adherence ranges in 1.4$\sim$2.0). Models such as MMAudio illustrate this pattern most clearly, producing audio that matches what is seen on screen but rarely incorporates the requested temporal, timbral, or intensity adjustments. The HunyuanVideo-Foley variants achieve slightly higher adherence but still remain far from reliable, underscoring that current architectures do not effectively integrate textual control signals. This visual bias ultimately limits controllable generation: when visual cues and textual instructions conflict, models overwhelmingly favor the visual stream, disregarding the user-specified behaviors.

\begin{table}[tb]
\centering
\small
\begin{tabular}{lcc}
\toprule
\textbf{Model} & \textbf{Recall} & \textbf{F1 Score} \\
\midrule
Qwen3-VL-8B-Instruct & 0.48 & 0.48 \\
Qwen3-VL-8B-Thinking & 0.44 & 0.50 \\
Qwen3-VL-30B-Instruct & 0.39 & 0.49 \\
Qwen3-VL-30B-Thinking & 0.45 & 0.53 \\
OmniVinci & 0.13 & 0.19 \\
Gemini-2.5 Flash & 0.66 & 0.50 \\
Gemini-2.5 Pro & 0.66 & 0.59 \\
Qwen3-Omni-30B-Instruct & 0.42 & 0.43 \\
Qwen3-Omni-30B-Thinking & 0.48 & 0.56 \\
\midrule
\textbf{Gemini-2.5 Pro + SF} & \textbf{0.83} $_{\textcolor{ForestGreen}{\bf +0.17}}$ & \textbf{0.74} $_{\textcolor{ForestGreen}{\bf +0.15}}$ \\
\textbf{Qwen3-VL-30B-Thinking + SF} & \textbf{0.54} $_{\textcolor{ForestGreen}{\bf +0.09}}$ & \textbf{0.71} $_{\textcolor{ForestGreen}{\bf +0.18}}$ \\
\bottomrule
\end{tabular}
\vspace{-0.3cm}
\caption{\textbf{Task 1: Sounding Event Detection.} Evaluating the ability of models to enumerate and correctly identify all sounding events in a video. “+SF” denotes our proposed Slow–Fast reasoning strategy. Results are averaged over three runs.}
\label{tab:sound_event_detection}
\vspace{-0.5cm}
\end{table}

\noindent\textbf{The Finer-grained Level, the Worse Control.}
As shown in Figure~\ref{fig:level}, controllability decreases sharply as the control level becomes more fine-grained. All models achieve their highest scores under video control, where the instruction specifies coarse-grained or video-level behavior. However, performance drops substantially for group control, and even further for individual control, which requires precise manipulation of individual sounding events. This widening gap reveals that existing VT2A models struggle to localize, disentangle, and manipulate fine-grained event attributes.

\subsection{Sounding Event Awareness}
\label{sec:sounding_event_awareness}

Beyond controllable audio generation, \BenchName{} also \textit{naturally} functions as a rigorous benchmark for evaluating a model’s \textbf{sounding event awareness}—its ability to detect and temporally localize sounding events in video. We design two complementary tasks to assess nine recent VideoLLMs and omni-modal foundation models.

\noindent\textbf{Task 1: Sounding Event Detection.} This task measures a model’s capacity to exhaustively identify all sounding events occurring within a video, reflecting its fundamental awareness of audio-relevant visual cues.
We evaluate nine models on 300 videos with human-verified sounding-event annotations with zero-shot, single-turn prompting protocol.
Table~\ref{tab:sound_event_detection} presents recall and F1 scores. Gemini-2.5 Pro demonstrates the highest performance. Omni-modal models generally outperform vision-centric VideoLLMs, indicating the importance of audio-aligned multimodal pretraining for event-level understanding.


\begin{figure}
    \centering
    \includegraphics[width=0.8\linewidth]{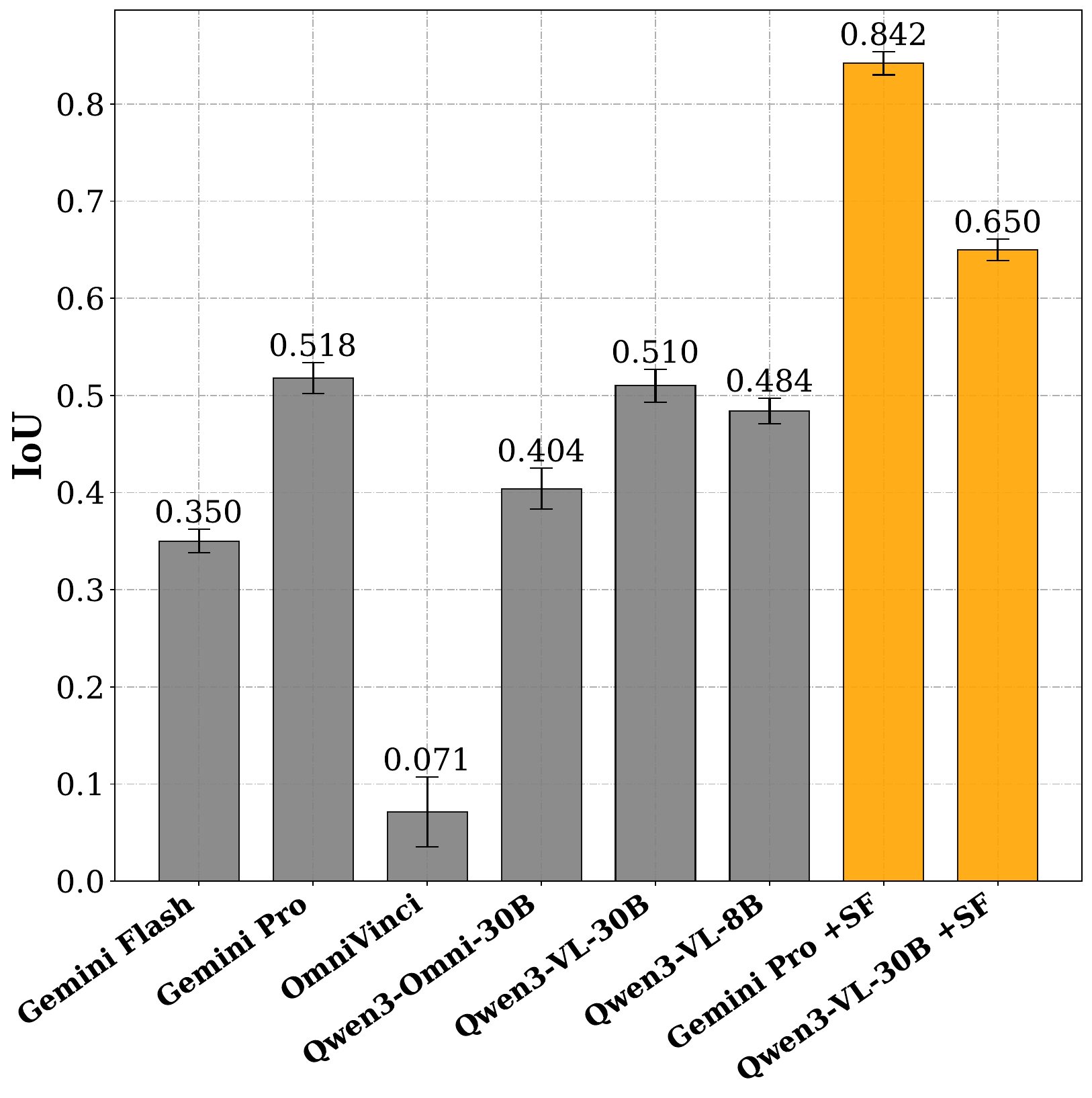}
    \vspace{-0.4cm}
     \caption{\textbf{Task 2: Sounding Event Localization.} 
    Intersection-over-Union between predicted and ground-truth temporal spans, computed only on correctly detected events. }
    \label{fig:timestamp_error}
    \vspace{-0.4cm}
\end{figure}

\noindent\textbf{Task 2: Sounding Event Localization.}
This task evaluates whether a model can not only detect sounding events but also accurately localize their temporal boundaries.
Using the same 300 videos, we assess temporal alignment only for correctly predicted events. For each such event, we compare the model’s predicted temporal span with the human-annotated ground truth and compute the IoU for measurement.
Figure~\ref{fig:timestamp_error} presents IoU scores across models. All models exhibit notable boundary drift—indicating that while they often detect \emph{which} events occur, they struggle to match the precise onset and offset times.

%% file: sec/06-method.tex
\section{Method: \textit{EchoVidia}}
\label{sec:method}

To address the limitations of existing video-to-audio generation models on lack of fine-grained controllability and vision dominance, we propose \ModelName{}, a training-free agentic framework with slow-fast thinking. In this section, we first introduces the design and core components of the proposed method (§\ref{sec:method_overview}), and then presents the experimental setup and results (§\ref{sec:method_evaluation}).

\subsection{Overview}
\label{sec:method_overview}

\noindent\textbf{Slow–Fast Thinking Strategy.}
As analyzed in Section~\ref{sec:sounding_event_awareness}, current VideoLLMs exhibit weak awareness of sounding events and struggle with accurate temporal grounding. 
Inspired by dual-process cognition where System 1 performs fast intuitive reasoning and System 2 conducts slow analytical reasoning, we first introduce a \emph{slow–fast thinking} strategy (SF) to enhance event understanding and timestamp localization. 
The \emph{fast thinking} pathway captures a global overview of the video in 1 fps, summarizing its high-level structure and coarse auditory context. 
The \emph{slow thinking} pathway performs detailed reasoning by 
viewing its 16x slower-motion video. Specifically, we first downsample the video to 16 fps, then temporally stretch it by 16x to obtain the video at 1 fps, enabling precise event localization and attribute inference. 

\begin{figure}[t]
    \centering
    \includegraphics[width=1\linewidth]{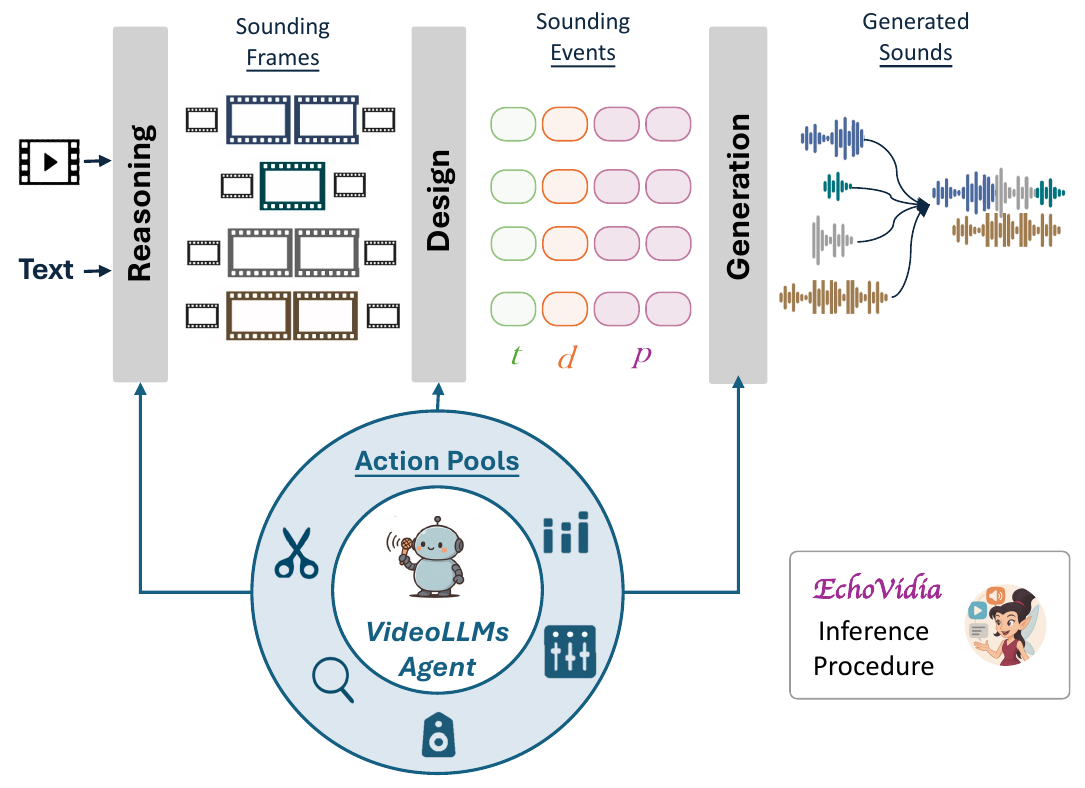}
    \vspace{-0.55cm}
    \caption{\ModelName{} Inference Procedure}
    \vspace{-0.5cm}
    \label{fig:method}
\end{figure}

\noindent\textbf{Architecture.}
The core of \ModelName{} is an agentic framework that structures the generation process into three interconnected stages: reasoning, design, and synthesis. 
At its center is a VideoLLM-based agent that interacts with an \emph{action pool}—a set of 12 atomic operations spanning visual reasoning and sound generation (see appendix for more details).
Within this pool, video reasoning actions identify sounding events, retrieve temporal cues, and crop relevant visual segments. 
Sound design actions allow the agent to add, remove, or modify event representations, controlling their semantic attributes and temporal properties. 
Finally, generation actions synthesize, adjust, and mix the resulting audio layers, ensuring both temporal alignment and perceptual coherence between modalities.

\noindent\textbf{Inference Procedure.}
As illustrated in Figure \ref{fig:method}, during inference, the agent first analyzes the input video to identify potential sounding events and estimate their approximate timing. 
It then constructs a symbolic event plan describing how each event should sound and iteratively refines this plan through reasoning and editing actions. 
The finalized symbolic representation is passed to the sound generation module, which renders the corresponding audio conditioned on both visual and textual contexts. 

\subsection{Results}
\label{sec:method_evaluation}
We compare our \ModelName{} with other video-text-to-audio models following the evaluation settings in Section~\ref{sec:experiment}. 

\noindent\textbf{Performance on \TaskName{}.} \ModelName{} achieves the best performance across most evaluation dimensions, outperforming every baseline model by a substantial margin, as shown in Table~\ref{tab:combined_results}. For controllability, \ModelName{} reaches 0.72 TemporalCtl, 0.78 TimbreCtl, and 0.75 VolCtl, corresponding to average improvements of roughly 55\% over the strongest baseline (outperforming the best baseline by +0.29, +0.30, and +0.19, respectively). Human evaluation further confirms the advantage: \ModelName{} attains 3.80 Instruction Adherence, 3.93 Audio–Visual Coherence, and 3.79 Perceptual Quality, consistently outperforming the strongest baseline by +1.20, +0.40, and +0.32, respectively.

Figure~\ref{fig:visualbias} illustrates that \ModelName{} achieves a \emph{balanced integration of visual and textual conditioning}.
Whereas existing models cluster in the upper-left region—showing reasonable Audio–Visual Coherence but poor Instruction Adherence—\ModelName{} uniquely attains simultaneously high scores on both axes. This shows that \ModelName{} not only synchronizes sound with visual content but also executes fine-grained textual controls faithfully, eliminating the visual-dominance bias observed in previous VT2A systems.

\noindent\textbf{Ablation on Slow–Fast (SF) Thinking Strategy.} As shown in Table~\ref{tab:sound_event_detection} and Figure~\ref{fig:timestamp_error}, models equipped with the SF reasoning strategy exhibit \textbf{substantial gains} in sounding-event awareness. For T1, SF boosts recall from 0.66 to 0.83 for Gemini-2.5 Pro and increases the F1 score from 0.59 to 0.74, representing the highest scores among all evaluated models. Similarly, Qwen3-VL-30B-Thinking benefits noticeably from SF, achieving a +0.15 gain in the F1 score. Beyond enumeration, SF also delivers strong gains in timestamp accuracy: for T2, the IoU of Gemini-2.5 Pro increases from 0.510 to 0.842, and Qwen3-VL-30B improves from 0.484 to 0.650, corresponding to more than a 60\% relative improvement in temporal precision.



%% file: sec/07-conclusion.tex
\section{Conclusion}
\label{sec:conclusion}
In conclusion, we introduced \TaskName{}, a new task focusing on fine-grained video-grounded sound generation with event-level hierarchical control. By shifting the focus from coarse video-level prompts to symbolic, temporally grounded sounding events, our formulation provides a principled way to specify \emph{what} sound should occur, \emph{when} it should occur, and \emph{how} it should evolve. To support systematic study of this task, we constructed \BenchName{}, a large-scale, expertly curated benchmark and metrics. We developed \ModelName{}, a training-free agentic framework with slow–fast thinking that significantly improves controllability, grounding fidelity, and perceptual quality. Future research may integrate our event-centric formulation into end-to-end trainable models, and extend the symbolic event representation to more application to further broaden the creative generation capabilities of generative model. We hope this work inspires future research toward omni-modal generative intelligence that can both understand and recreate the multimodal richness of the real world.

%% file: sec/suppl.tex
\clearpage
\setcounter{page}{1}

\appendix
\appendixtableofcontents
\clearpage

\input{supp/A-benchmark}
\input{supp/B-method}
\input{supp/C-case_study}


%% file: supp/A-benchmark.tex
\section{Border Impact}

Our proposed framework opens up new opportunities for controllable and interpretable VT2A generation.  By enabling fine-grained instruction-guided sound control, it can benefit multiple downstream applications, including video editing, film post-production, and accessible multimedia creation.
The data curation pipeline for \BenchName{} and the \ModelName{} model together provide a scalable data generation pipeline for synthesizing high-quality sounding videos with rich, aligned multimodal annotations.
Such controllable generation pipelines not only enhance content creation but also serve as a valuable source of training data for large-scale \emph{world models} and \emph{omni-modal foundation models}, which aim to unify perception, reasoning, and generation across modalities.

\section{Details on Benchmark: EchoFoley-6k}

\subsection{ Data Example}

Each data point in EchoFoley-6k follows the triplet format 
$\langle$video, instruction, sounding-event annotations$\rangle$.

\begin{itemize}
    \item \textbf{Video}: a motion-centered clip (6--30 seconds) where sound-producing interactions
    are visually evident.
    \item \textbf{Instruction}: a natural-language directive indicating how sounds should be 
    generated, edited, or transformed. Instructions may specify instance-, group-, or video-level 
    control (e.g., ``change the second meow into a lion roar'' or ``make all prior sounds louder'').
    \item \textbf{Sounding-Event Annotations}: a set of symbolic sounding events
    $e = (t, d, p)$, where $t = (t_{\text{start}}, t_{\text{end}})$ denotes the temporal span,
    $d$ is a semantic description, and $p$ contains controllable auditory attributes
    (pitch, timbre, loudness, intensity, spatialization).
\end{itemize}

Figure~1 in the main paper shows an example illustrating how temporal boundaries,
semantic descriptions, and controllable properties interact with instructions.

\subsection{ Comparison with Existing Datasets}

Existing audio--visual datasets primarily support coarse audio modeling and do not provide 
fine-grained event boundaries or hierarchical controllability. 
EchoFoley-6k fills this gap by jointly providing:
\begin{itemize}
    \item dense event-level temporal boundaries,
    \item symbolic, multi-attribute sounding-event representations,
    \item natural-language instructions for hierarchical control.
\end{itemize}

\begin{table}[h]
\centering
\small
\begin{tabular}{lcc|cccc}
\toprule
\textbf{Dataset} & 
\multicolumn{2}{c|}{\textbf{Data}} &
\multicolumn{4}{c}{\textbf{Annotation}} \\
\cmidrule(lr){2-3} \cmidrule(lr){4-7}
& \textbf{Video} & \textbf{Instruction} &
\textbf{Audio Desc.} &
\textbf{Audio Temp. Cap.} &
\textbf{Audio Vol. Cap.} &
\textbf{Event-Centric Ann.} \\
\midrule
VGGSound \cite{vggsound}              & \textbf{Yes} & No  & \textbf{Yes} & No  & No  & No \\
Ego4Dsounds \cite{ego4Dsounds}        & \textbf{Yes} & No  & \textbf{Yes} & No  & No  & No \\
AudioSet \cite{audioset}              & No  & No  & \textbf{Yes} & No  & No  & No \\
AudioCaps \cite{audiocaps}            & No  & No  & \textbf{Yes} & No  & No  & No \\
WavCaps \cite{wavcaps}                & No  & No  & \textbf{Yes} & No  & No  & No \\
BBC Sound Effects \cite{BBCSoundEffects} & No  & No  & \textbf{Yes} & No  & No  & No \\
AudioCoT \cite{liu2025thinksound}     & \textbf{Yes} & No  & \textbf{Yes} & No  & No  & No \\
Kling-Audio-Eval \cite{wang2025kling} & \textbf{Yes} & No  & \textbf{Yes} & No  & No  & No \\
MovieGen Bench \cite{moviegen}        & \textbf{Yes} & No  & \textbf{Yes} & No  & No  & No \\
AVVP \cite{tian2020avvp}              & \textbf{Yes} & No  & \textbf{Yes} & \textbf{Yes} & No  & No \\
ASVA \cite{linz2024asva}              & \textbf{Yes} & No  & \textbf{Yes} & \textbf{Yes} & No  & No \\
MA-Bench \cite{rong2025audiogenie}    & \textbf{Yes} & No  & \textbf{Yes} & \textbf{Yes} & \textbf{Yes} & No \\
\midrule
\BenchName{}                          & \textbf{Yes} & \textbf{Yes} & \textbf{Yes} & \textbf{Yes} & \textbf{Yes} & \textbf{Yes} \\
\bottomrule
\end{tabular}
\caption{Comparison of datasets and available annotation types.}
\end{table}

\subsection{ Data Curation Details}

The full data construction pipeline is illustrated in Figure~2 of the main paper. Below we describe each stage in detail.

\subsubsection{Curation Process Details} 

\begin{enumerate}
    \item \textbf{Video Filtering}.  
    We begin with motion-centered videos containing visually identifiable sound-producing actions 
    (e.g., collisions, vocalizations, material interactions). Videos with ambiguous, off-screen, or purely ambient sounds are removed to ensure that sounding events remain visually grounded.

    \item \textbf{Metadata and Frame Captioning}.  
    For each candidate video, we extract its metadata (title and textual description) and generate frame-level visual captions at 16~fps using Gemini 2.5 pro\cite{gemini}.  
    These captions summarize evolving object configurations, motions, and interactions, providing structured visual grounding for downstream event extraction and reasoning.

    \item \textbf{Story Proposal and Event Extraction}.  
    Using GPT-5\cite{gpt5}, we generate imaginative narratives describing plausible auditory interpretations of the scene.  
    The model also proposes an initial list of hypothesized sounding events, each with an approximate temporal region and a short semantic description.  
    These machine-generated proposals serve as high-level scaffolds and are not treated as final annotations.

    \item \textbf{Human Refinement}.  
    Human annotators transform the model-generated narratives into executable, fine-grained instructions; refine all event boundaries using frame-by-frame inspection; remove any hallucinated or visually unsupported events; and annotate rich, multi-attribute sound properties 
    (e.g., pitch, timbre, volume, intensity, spatialization).  
    This stage yields temporally accurate and instruction-aligned sounding-event annotations of high quality.

\end{enumerate}

\subsubsection{Human Refinement Instruction}

\noindent\textbf{Context.}
Human annotators verify and refine AI-generated instructions and their corresponding manipulated sound annotations. The goal is to ensure that the final dataset is logically coherent, unambiguous, and free of AI errors.

\medskip
\noindent\textbf{General Principles.}
Annotators are provided with a silent video, an AI-generated instruction, and the corresponding manipulated annotation file. They are asked to validate the following three aspects:

\begin{itemize}
    \item \textit{Logical Coherence.}
    The instruction must be plausible within the visual world of the video. The described sound should be something that could reasonably occur in the depicted environment, even if it is not currently present.
    The key question is: \emph{Could this sound plausibly occur in this scene?}

    \item \textit{Instructional Clarity.}
    The instruction must be specific and unambiguous, allowing an annotator to execute it without guessing the intended operation.
    The key question is: \emph{Do I know exactly what action to take without having to infer missing details?}

    \item \textit{Annotation Accuracy.}
    The manipulated annotation file must be an exact execution of the validated instruction, with no inconsistencies in timing, wording, or structure.
    The key question is: \emph{Is the final annotation a perfect realization of the instruction, with zero errors?}
\end{itemize}

\medskip
\noindent\textbf{Core Annotation Tasks.}

\noindent\textit{T1. Verify the Instruction.}
Annotators first decide whether the instruction is both plausible and clear:
\begin{itemize}
    \item \textbf{Plausibility.} Annotators watch the video and check whether the instruction is logically grounded in the scene. Instructions that cannot be reconciled with the visual context (e.g., adding an animal sound to a scene with no animals or outdoor context) are rejected.
    \item \textbf{Clarity.} Annotators read the instruction and determine whether it is precise enough to be executed. Instructions that are vague or subjective (e.g., ``make the sound more exciting'') are flagged, with a brief comment explaining the source of ambiguity.
\end{itemize}

\noindent\textit{T2. Verify and Correct the Annotation.}
If the instruction passes T1, annotators then meticulously verify and correct the AI-generated manipulated annotation:
\begin{itemize}
    \item \textbf{Timestamps.} For instructions that shift or rescale time (e.g., ``start 1.5 seconds earlier''), annotators recompute the new timestamps themselves (e.g., 00:05.000 $\rightarrow$ 00:03.500) and confirm that all start and end times are exact and consistent.
    \item \textbf{Descriptions.} When textual descriptions are modified, annotators check that the changes are integrated fluently (e.g., adding ``metallic'' yields ``a metallic chirp'' rather than ungrammatical phrasing) and that the semantics remain faithful to the instruction.
    \item \textbf{Structure.} When events are added, removed, or reordered, annotators scan the full event list to ensure that only the intended events have been modified, and that no unrelated events have been accidentally altered or deleted.
\end{itemize}

This procedure ensures that every example in the dataset results from a valid, human-verified instruction and a perfectly aligned manipulated annotation.

\subsection{Automatic Evaluation Metrics Implementation Details}

\subsubsection{Temporal Control}

To evaluate how accurately the generated audio follows the intended timing of each sounding event, we measure temporal controllability by comparing the annotated event interval with the event's predicted start and end times extracted from the generated audio.

\textbf{Audio preprocessing.}
The generated waveform is resampled to 16~kHz and converted into a 64-bin log-mel spectrogram using 25~ms Hann windows with a 10~ms hop size. The spectrogram is normalized and used as input to the temporal localization module.

\textbf{Gemini-based onset/offset prediction.}
For temporal localization, we directly query a Gemini2.5-pro\cite{gemini} with the generated audio segment and the textual description of the event. Gemini is prompted to determine when the described sound first appears and when it ends, returning a predicted start and end time in seconds. We use the following prompt to guide Gemini in predicting the temporal boundaries of each event:

\begin{tcolorbox}
\begin{footnotesize}
\begin{verbatim}
You are an expert audio analyst. 
Given an audio clip and a textual description of a sound event, 
identify when this event starts and when it ends.

The event description is:
"{EVENT_DESCRIPTION}"

Please listen to the audio and return:
- The start time of this event (in seconds)
- The end time of this event (in seconds)

If the event occurs multiple times, choose the occurrence 
that best matches the description. If unsure, choose the 
most prominent occurrence.

Output your answer in the following JSON format:
{
  "start_time": <float, seconds>,
  "end_time": <float, seconds>
}
\end{verbatim}
\end{footnotesize}
\end{tcolorbox}

\textbf{Interval overlap estimation.}
The predicted interval is compared with the annotated interval to assess alignment. Strong overlap indicates precise temporal localization, while little or no overlap reflects temporal drift. The temporal controllability score for a video is computed by averaging the alignment scores across all instruction-relevant events.

\begin{tcolorbox}
\begin{footnotesize}
\begin{verbatim}
Pseudocode for Temporal Controllability (TempCtl)

Input:
    - Ground-truth events C
    - For each event: annotated timestamp
    - Generated audio A_hat
    - Gemini-based boundary predictor D_time

Procedure:
    score = 0
    For each event e in C:
        # Ask Gemini to predict event boundaries
        t_pred = D_time(A_hat, e.description)

        # Compute overlap with ground truth
        inter_len = intersection_length(annotated(e), t_pred)
        union_len = union_length(annotated(e), t_pred)

        if union_len > 0:
            temp_score = inter_len / union_len
        else:
            temp_score = 0

        score += temp_score

Output:
    TempCtl = score / |C|
\end{verbatim}
\end{footnotesize}
\end{tcolorbox}

\subsubsection{Timbre Control}

Timbre controllability measures whether the generated audio segment for each event matches the intended sound identity specified by the instruction.

\textbf{Audio cropping.}
For each event, the annotated start and end times are converted into sample indices and used to extract the corresponding waveform segment. If the segment is longer than the analysis window, a sliding window strategy is used; if shorter, zero-padding is applied.

\textbf{CLAP-based semantic alignment.}
We employ the audio--text encoder of CLAP\cite{laionclap2023} to compute timbre similarity. The cropped audio is resampled to 48~kHz, converted to mono, amplitude-normalized, and fed to CLAP's audio encoder. The event description is processed by the CLAP text encoder. Both encoders output normalized embeddings, and a cosine similarity score reflects how well the generated sound matches the desired auditory identity.

\textbf{Aggregation.}
Scores across all events described or modified by the instruction are averaged to produce the timbre controllability score.

\begin{tcolorbox}
\begin{footnotesize}
\begin{verbatim}
Pseudocode for Timbre Controllability (TimbCtl)

Inputs:
    - Event set C
    - For each event e in C:
        * e.start_time, e.end_time (in seconds)
        * e.description (target timbre text)
    - Generated audio A_hat with sampling rate sr
    - CLAP audio encoder CLAP_audio(·)
    - CLAP text encoder  CLAP_text(·)
    - Fixed analysis length L samples

Procedure:
    total_score = 0
    valid_events = 0

    for each event e in C:

        # 1. Time to sample indices
        s = floor(e.start_time * sr)
        t = ceil(e.end_time  * sr)
        s = clamp(s, 0, length(A_hat)-1)
        t = clamp(t, s+1, length(A_hat))

        # 2. Crop and adjust segment length
        seg = A_hat[s : t]
        seg = pad_or_trim(seg, L)       # zero-pad or center-crop to L

        # 3. CLAP preprocessing + encoding
        seg_proc = CLAP_preprocess(seg) # resample, mono, normalize
        a_emb = normalize( CLAP_audio(seg_proc) )
        t_emb = normalize( CLAP_text(e.description) )

        # 4. Cosine similarity for this event
        sim = cosine(a_emb, t_emb)
        total_score += sim
        valid_events += 1

    if valid_events > 0:
        TimbCtl = total_score / valid_events
    else:
        TimbCtl = 0

Output:
    TimbCtl
\end{verbatim}
\end{footnotesize}
\end{tcolorbox}

\subsubsection{Volume Control}

Volume controllability evaluates whether the generated audio reflects the intended loudness pattern for each sounding event, relative to the global loudness of the track.

\textbf{Perceptual loudness extraction.}
We compute loudness using a perceptual RMS measure based on the ITU-R BS.1770 K-weighting filter. This produces a perceptually grounded measure for both the full generated audio and each event segment.

\textbf{Relative loudness classification.}
To make the evaluation invariant to global gain differences, each event's loudness is normalized by the global loudness of the audio. The resulting ratio is mapped into ``low'', ``medium'', or ``high'' categories using two thresholds calibrated on the development set.

\textbf{Scoring.}
A prediction is correct if the generated loudness category matches the annotated label. The final score is the proportion of correctly classified events.

\begin{tcolorbox}
\begin{footnotesize}
\begin{verbatim}
Pseudocode for Volume Controllability (VolCtl)

Input:
    - Event set C
    - For each event: timestamp, loudness label
    - Generated audio A_hat
    - Loudness(): perceptual RMS function
    - Thresholds tau1, tau2

Procedure:
    global_L = Loudness(A_hat)

    correct = 0
    For each event e in C:
        seg = crop_audio(A_hat, e.timestamp)
        seg_L = Loudness(seg)
        r = seg_L / global_L

        if r < tau1:      pred = "low"
        elif r < tau2:    pred = "medium"
        else:             pred = "high"

        if pred == e.loudness_label:
            correct += 1

Output:
    VolCtl = correct / |C|
\end{verbatim}
\end{footnotesize}
\end{tcolorbox}

\subsection{Human Evaluation Details}

\subsubsection{Survey Design}

\begin{figure}[h]
    \centering
    \includegraphics[width=1\linewidth]{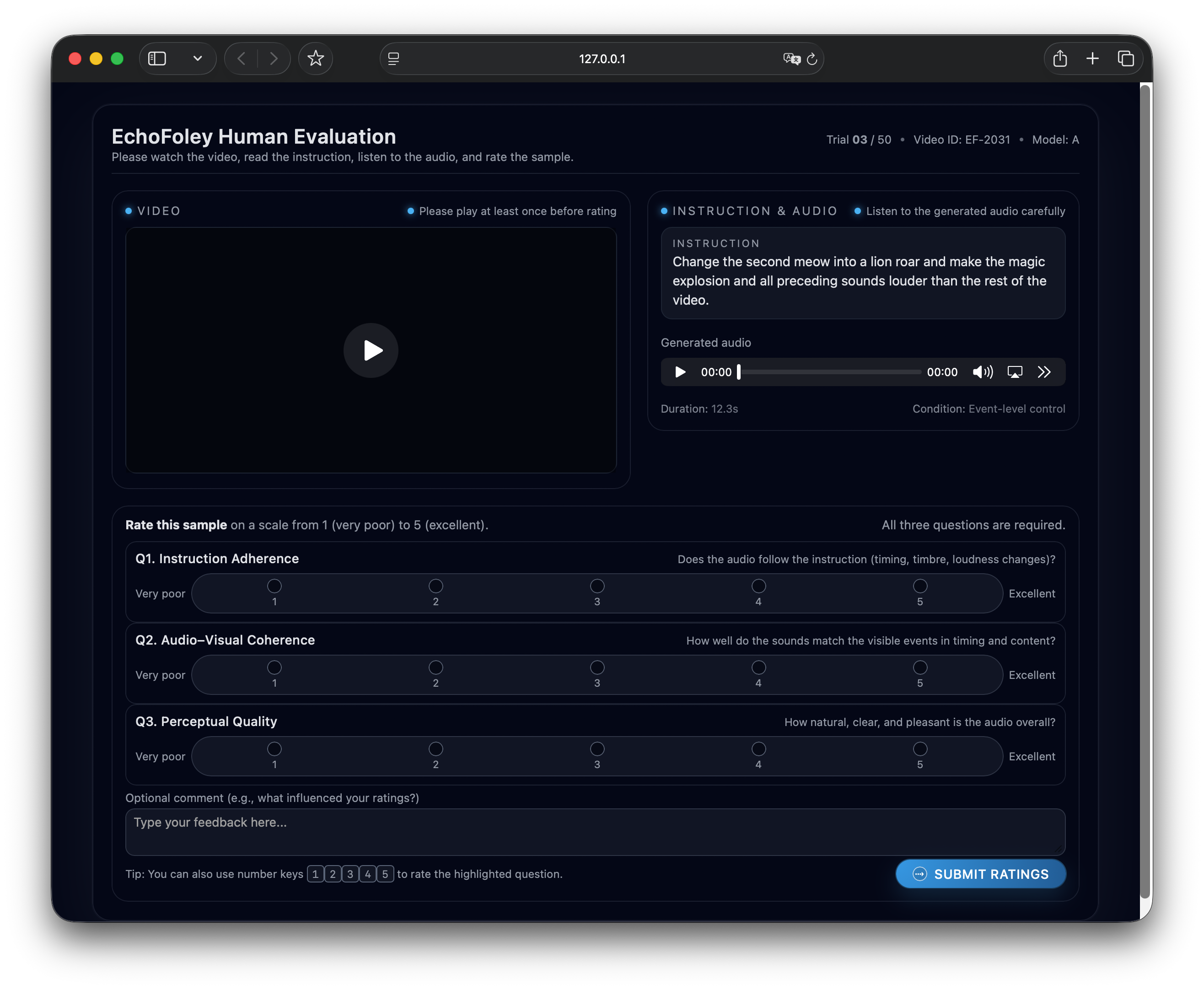}
    \caption{Human Evaluation UI}
    \label{fig:huam_eval_ui}
\end{figure}

We design a three-part human evaluation to assess perceptual aspects of controllable
video-to-audio generation that are difficult to capture algorithmically. 
Annotators rate each generated audio clip on a 1--5 Likert scale along the following
dimensions:

\begin{itemize}
    \item \textbf{Instruction Adherence:} 
    How well the generated audio follows the user's instruction, including
    requested changes in timing, timbre, or loudness.

    \item \textbf{Audio--Visual Coherence:} 
    How consistent the audio is with the visual content, including synchronization
    with object actions, motion dynamics, and event boundaries.

    \item \textbf{Perceptual Quality:} 
    The overall naturalness, clarity, and realism of the generated audio
    within the video context.
\end{itemize}

Annotators are also provided with the original silent video, the instruction, 
and the generated audio to ensure consistent evaluation conditions.

\subsubsection{Human Evaluation Procedure}

We randomly sample 50 video--instruction pairs from EchoFoley-6k and
generate outputs from all evaluated models. Six annotators independently rate
each audio clip using the survey described above. To minimize potential bias,
annotators are not informed of the identity of the model that produced each clip,
and all clips are presented in randomized order. 

For each metric, we compute the average score across annotators and clips.
Inter-annotator agreement is measured using Cohen's kappa, resulting in a value
of 0.62, indicating substantial agreement across raters. This procedure provides
a reliable perceptual assessment complementing the automatic controllability metrics.

\subsection{Evaluation Experiment Setup Details}
For sound generation evaluation, we random choose 100 cases from \BenchName{} for evaluation. The baseline model configuration remain default for all models. Note that since AudioGenie\cite{rong2025audiogenie} is a huge frameowrk contains irrelevant packaged and models for other tasks ( e.g. Text-to-Music, Text-to-Speech, etc.), for the sake of depolyment efficency, we implement the minimal version with only Video-Text-To-Audio functions avalibile, the remaining structure remain unchanged.

For sounding event awareness evaluation, we run 3 times for all unique videos in our dataset for each model, and take the average. We use all the default parameters (e.g. temprature, top-p, etc.) for the model.

%% file: supp/B-method.tex
\section{Details on Method: \textit{EchoVidia}}
\label{appendix:method}

This section provides additional details of the proposed \textit{EchoVidia} framework, including the slow–fast thinking strategy, action-pool architecture, implementation details, and limitations.

\subsection{Slow--Fast Thinking Strategy Details}
\label{appendix:sf}

The Slow–Fast (SF) Thinking strategy is designed to compensate for the limited sounding-event awareness found in current VideoLLMs. As detailed in Sec.~6 of the main paper, SF integrates two complementary temporal reasoning pathways:

\paragraph{Fast Thinking (Global View).}
We extract a 1~fps version of the video, preserving only coarse temporal structure. This compressed view encourages the VideoLLM to capture global scene dynamics, high-level semantic context, and broad sequencing of potential sounding events. These global cues help the model form initial hypotheses regarding (1) what categories of events may occur, (2) their approximate ordering, and (3) the overall auditory context (e.g., repetitive actions, scene transitions).

\paragraph{Slow Thinking (Fine-Grained View).}
To support precise timestamp localization, we downsample the input video to 16~fps and then temporally stretch it by a factor of 16$\times$. This effectively presents the model with an ultra-slow-motion view, enabling accurate inspection of subtle motions (e.g., object impacts, mouth articulations, or momentary gestures) that correspond to sounding events. The slow-stream reasoning significantly improves the detection of event boundaries and facilitates fine-grained attribute inference (e.g., intensity, pitch proxy, or spatial clues).

\paragraph{Integration.}
The VideoLLM processes both views independently. We aggregate their outputs by:
(a) merging candidate sounding events;
(b) reconciling coarse timestamps with fine-grained slow-view refinements; and  
(c) resolving inconsistencies by prioritizing slow-view boundaries when conflict arises.
This integration forms the event plan used in the symbolic representation.

\subsection{Framework Architecture Details}
\label{appendix:architecture}

EchoVidia is implemented as an agentic multi-stage pipeline in which a VideoLLM-based controller sequentially invokes a set of atomic operations, referred to as the \textit{action pool}. Instead of relying on a monolithic forward pass, EchoVidia decomposes video-to-audio generation into three explicit phases—\textit{reasoning}, \textit{sound design}, and \textit{synthesis}. Each phase is realized through modular actions with well-defined inputs, outputs, and side effects. This decomposition enables the agent to iteratively refine symbolic sounding-event representations, perform targeted corrections, and maintain global temporal consistency.

\begin{table}[h]
\centering
\small
\begin{tabular}{p{3.6cm}p{9.4cm}}
\toprule
\textbf{Action Category} & \textbf{Description} \\
\midrule
\multicolumn{2}{l}{\textbf{Video Reasoning Actions}} \\
\midrule
Query sounding event & Retrieve sounding events based on a semantic query (e.g., ``the second meow''). \\
Query timestamp & Estimate onset/offset timestamps or refine event boundaries. \\
Crop video footage & Extract a subclip for localized inspection during slow–fast reasoning. \\
Adjust video speed & Present the video at altered framerates (slow-motion or accelerated) to improve temporal precision. \\
\midrule
\multicolumn{2}{l}{\textbf{Sound Generation Actions}} \\
\midrule
Generate audio & Synthesize an audio segment given its symbolic description and temporal span. \\
Tune audio volume & Adjust loudness or relative gain for a target event. \\
Mix audio tracks & Merge event-level audio layers with crossfading and temporal alignment. \\
\midrule
\multicolumn{2}{l}{\textbf{Sound Design Actions}} \\
\midrule
Add event & Insert a new sounding event into the event plan. \\
Delete event & Remove an existing event from the symbolic representation. \\
Modify event description & Change semantic attributes (e.g., ``cat meow'' $\rightarrow$ ``lion roar''). \\
Modify event time & Adjust event timestamps or duration while preserving ordering constraints. \\
Modify event properties & Edit timbre or perceptual attributes (volume, pitch, intensity, spatial width). \\
\bottomrule
\end{tabular}
\caption{Action pool used by the \ModelName{} agent.}
\label{tab:action_pool}
\end{table}

\paragraph{Action Abstractions.}
EchoVidia treats each action in the pool as a callable transformation governed by a standardized interface:

\begin{itemize}
    \item \textbf{Input:} a structured state object consisting of the video clip, optional cropped subclips, the current symbolic event plan, and the user instruction.
    \item \textbf{Operation:} an atomic transformation that updates either (i) the \textit{event latent state} (i.e., symbolic representation), or (ii) the \textit{audio latent state} (i.e., generator instructions), or (iii) the \textit{visual latent state} (i.e., crop or resample metadata).
    \item \textbf{Output:} an updated state object that becomes the input for the next action.
\end{itemize}

This design enables multi-step planning, rollback of intermediate errors, and flexible recomposition of operations depending on the complexity of the instruction.

\paragraph{Action Pool.}
Table~\ref{tab:action_pool} summarizes the full set of actions used by the agent. Video reasoning actions support fine-grained temporal grounding and visual context extraction; sound-design actions manipulate the symbolic representation of sounding events; and sound-generation actions interface with the audio synthesis backend. The modular nature of the action pool allows the agent to construct arbitrarily complex behaviors through few-shot prompting rather than bespoke training.

\paragraph{Execution Flow.}
During inference, the agent begins with a high-level reasoning pass in which it identifies sounding events, estimates their temporal structure, and formulates an initial symbolic plan. It then repeatedly applies sound-design actions to modify the event plan in accordance with user instructions, ensuring correct temporal alignment and attribute-level control. Finally, the agent invokes generation actions to render each event as audio and perform mixing operations to synthesize the final output waveform.

\subsection{Implementation Details}
\label{appendix:implementation}

\subsubsection{Prompt Template}

We design a structured prompt template that guides the VideoLLM agent for different stage.

\noindent\textbf{Video Overview Prompt:} generating a dense timestamped log of diegetic sound events.
\begin{tcolorbox}
\begin{footnotesize}
\begin{verbatim}
SYSTEM ROLE:
You are a Video-to-Sound Inference Specialist. You analyze a silent video
chronologically and infer all plausible diegetic sound events caused by
actions, motions, or interactions visible in the scene.

INPUT:
<VIDEO_FRAMES>

TASK:
Identify every visually implied non-speech, non-ambient sound.
For each sound event, infer the most specific and self-contained
description possible (e.g., "rubber sole scraping on tile floor"
instead of "footstep").
If multiple sounds occur concurrently, list each separately but share
the same timestamp.
Break repeated or interrupted sounds into separate timestamp entries.
Include descriptive attributes of sound quality (e.g., "abrupt",
"distant", "metallic", "soft").
Ignore background ambience, room tone, or non-action-based noise.

OUTPUT FORMAT (STRICT):
A flat list of entries in the format:
[TIMESTAMP] - [self-contained sound description]
EXAMPLE:
[00:02] - A heavy wooden door creaks open slowly.
[00:04] - Soft footsteps on a polished floor.
[00:07] - The sharp click of a light switch.
[00:07] - The low hum of a fluorescent light turning on.
[00:12] - A piece of paper rustles as it's picked up.
Do NOT provide any commentary, explanations, or additional text.
Only output the timestamped list.
\end{verbatim}
\end{footnotesize}
\end{tcolorbox}

\noindent\textbf{Video Event Detection Prompt:} summarizing global scene information and requesting enumeration of potential sounding events.
\begin{tcolorbox}
\begin{footnotesize}
\begin{verbatim}
SYSTEM ROLE:
You are a Video Understanding Specialist. You analyze visual content
and extract only information relevant to sound-producing events.

INPUT:
<VIDEO_FRAMES_1FPS>

TASK:
1. Describe the global scene (1–2 sentences).
2. Enumerate all visually identifiable actions that can produce sound.
3. For each action, provide:
   - a short semantic label (e.g., "cat meow", "object impact")
   - the rough order index (1, 2, 3, ...) WITHOUT timestamps
   - a short justification (why it may produce sound)

OUTPUT FORMAT (STRICT):
EVENT_LIST = [
  {index: i, label: "...", justification: "..."},
  ...
]
Do NOT include timestamps or any speculation unrelated to visible motion.
\end{verbatim}
\end{footnotesize}
\end{tcolorbox}

\noindent\textbf{Fine-grained Event Start-Time Localization Prompt:} pinpointing the exact onset of a sounding event within a local temporal window.
\begin{tcolorbox}
\begin{footnotesize}
\begin{verbatim}
SYSTEM ROLE:
You are a Video Analysis Expert. You analyze a silent video to pinpoint
the precise start time of a specific event, given a descriptive prompt
and an approximate timestamp as a search hint.
INPUT:
VIDEO: <VIDEO_FRAMES>
EVENT_DESCRIPTION: "<short natural-language description of the event>"
APPROX_TIMESTAMP: "MM:SS" (rough estimate of when the event occurs)
TASK:
Establish a Search Window:
Focus on the segment from 10 seconds before to 10 seconds after
the APPROX_TIMESTAMP.
Infer Key Visual Cues:
From EVENT_DESCRIPTION, determine the concrete physical action
that generates the sound (e.g., for "a meow": the initial opening
of the cat's mouth; for "a thud": the exact moment of impact).
Scan for the First Onset:
Within the search window, examine frames chronologically to find the
very first frame where the key visual cue for the event begins.
Decide the Precise Start Time:
The timestamp of this first onset frame is the precise start time
of the event.
Explain Your Decision:
Provide a short justification describing the visual evidence and how
it supports the chosen timestamp.
OUTPUT FORMAT (STRICT):
START_TIME_RESULT = {
timestamp: "MM:SS",
justification: "..."
}
Do NOT output multiple timestamps, and do NOT describe audio properties.
Focus strictly on visual evidence for the onset of the event.
\end{verbatim}
\end{footnotesize}
\end{tcolorbox}
\vspace{0.5em}
\noindent\textbf{Fine-grained Event End-Time Localization Prompt:} detecting the visual cue that signals the completion of an event.
\begin{tcolorbox}
\begin{footnotesize}
\begin{verbatim}
SYSTEM ROLE:
You are a Video Analysis Expert. You determine the precise end time of
a specific event in a silent video using a two-step reasoning process.
INPUT:
VIDEO: <VIDEO_FRAMES>
EVENT_DESCRIPTION: "<short natural-language description of the event>"
START_SEARCH_TIME: "MM:SS" (the time from which to begin scanning)
TASK:
Define the End Cue:
From EVENT_DESCRIPTION, infer the concrete visual action that marks
the completion of the event. This is the "End Cue".
Examples:
For "a meow": End Cue = "the cat's mouth is fully closed".
For "a door opening": End Cue = "the door stops moving".
Locate the First Completion of the End Cue:
Starting from START_SEARCH_TIME, scan frames chronologically and
identify the first moment at which the End Cue is fully satisfied.
The timestamp of that frame is the precise end time.
Provide Reasoning:
Briefly explain how the End Cue was defined and how it was visually
located in the video.
OUTPUT FORMAT (STRICT):
End Sign Description: <description of the visual End Cue>
Reasoning on how to locate the end time: <short reasoning>
End Time: MM:SS
Do NOT output any additional text, commentary, or multiple candidate
timestamps. Provide exactly one end time based solely on visual evidence.
\end{verbatim}
\end{footnotesize}
\end{tcolorbox}

\noindent\textbf{Slow–Fast Reasoning Fusion Prompt:} asking the model to reconcile global and fine-grained predictions.
\begin{tcolorbox}
\begin{footnotesize}
\begin{verbatim}
SYSTEM ROLE:
You are a Temporal Fusion Expert. Your job is to merge two event
streams into one consistent timeline.

INPUT:
FAST_VIEW_EVENTS:
<LIST_FROM_1FPS>

SLOW_VIEW_EVENTS:
<LIST_FROM_SLOW_MOTION>

TASK:
1. Merge events with similar semantics from both lists.
2. Refine event timing using SLOW_VIEW when available.
3. Assign timestamps (t_start, t_end) in seconds.
4. Remove duplicates and ensure chronological ordering.

OUTPUT FORMAT (STRICT):
MERGED_EVENTS = [
  {label: "...", t_start: x.xx, t_end: y.yy},
  ...
]
No explanations. Only the list above.
\end{verbatim}
\end{footnotesize}
\end{tcolorbox}

\noindent\textbf{Verification:} re-query timestamps or re-evaluate event descriptions if inconsistencies or contradictions arise.
\begin{tcolorbox}
\begin{footnotesize}
\begin{verbatim}
SYSTEM ROLE:
You are the Sounding Event Structuring Agent. You convert events
into symbolic representations for controllable audio generation.

INPUT:
MERGED_EVENTS:
<list from fusion step>

TASK:
For each event, construct e = (t, d, p):
 - t = (t_start, t_end)
 - d = {subject: ?, action: ?, object: optional}
 - p = {pitch: ?, volume: ?, intensity: ?, spatial: ?}

Rules:
- Infer d from visual cues only.
- Use DEFAULT for p attributes if uncertain.
- All fields must exist.

OUTPUT FORMAT (STRICT):
EVENT_PLAN = [
  {
    t: (x.xx, y.yy),
    d: {subject: "...", action: "...", object: "..."},
    p: {pitch: "...", volume: "...", intensity: "...", spatial: "..."}
  },
  ...
]
No free-form text. Only structured results.
\end{verbatim}
\end{footnotesize}
\end{tcolorbox}

\noindent\textbf{Editing Prompt.} applying user instructions for event insertion, modification, or deletion
\begin{tcolorbox}
\begin{footnotesize}
\begin{verbatim}
SYSTEM ROLE:
You are the Sound Design Controller. You update an existing symbolic
event plan based on user instructions.

INPUT:
USER_INSTRUCTION:
"<instruction text>"

CURRENT_EVENT_PLAN:
<symbolic plan>

TASK:
1. Identify all referenced events.
2. Apply the required edits using ONLY:
   - ADD_EVENT
   - DELETE_EVENT
   - MODIFY_DESCRIPTION
   - MODIFY_TIME
   - MODIFY_PROPERTIES
3. Validate chronological ordering and value ranges.

OUTPUT FORMAT (STRICT):
UPDATED_EVENT_PLAN = [
  {t: (...), d: {...}, p: {...}},
  ...
]
No reasoning statements. Only the updated plan.
\end{verbatim}
\end{footnotesize}
\end{tcolorbox}

\noindent\textbf{Generation Prompt:} specifying how the symbolic plan should be converted to audio-generation commands.

\begin{tcolorbox}
\begin{footnotesize}
\begin{verbatim}
SYSTEM ROLE:
You are the Audio Generation Planner. You convert symbolic events
into commands for the audio synthesis backend.

INPUT:
FINAL_EVENT_PLAN:
<symbolic plan>

TASK:
For each event, produce a generator command block with:
- event_id
- synthesis_prompt (derived from d and p)
- t_start / t_end
- acoustic properties

Then produce mixing instructions specifying:
- layering
- crossfades
- loudness normalization
- global effects

OUTPUT FORMAT (STRICT):
GENERATION_COMMANDS = [
  {
    event_id: i,
    synthesis_prompt: "<text prompt>",
    t_start: x.xx,
    t_end: y.yy,
    properties: {volume: ..., pitch: ..., intensity: ..., spatial: ...}
  },
  ...
]

MIXING_INSTRUCTIONS = {
  layering: "...",
  crossfade: "...",
  loudness_normalization: "...",
  global_effects: "..."
}
Only the structures above. No additional commentary.
\end{verbatim}
\end{footnotesize}
\end{tcolorbox}

\subsubsection{Base Models}

\paragraph{VideoLLMs.}
We use Gemini 2.5 Pro \cite{gemini} as the base model VideoLLM for event reasoning and audio planing. We combine its strong temporal understanding and multimodal grounding with our SF thinking strategy to enhance the performance. No additional fine-tuning is applied.

\paragraph{Audio Diffusion Model.}
We use Stable Audio Open 1.0 \cite{evans2025stable} as the base diffusion model for sound synthesis, leveraging its prompt-conditioning and high dynamic range. Event-level segments are generated individually and mixed with moviepy package.

\subsubsection{Computation Resources Requirement}

EchoVidia is entirely training-free. All evaluations were conducted using:
\begin{itemize}
    \item \textbf{GPU:} A single NVIDIA A100 (80GB).  
    \item \textbf{CPU:} 32-core Xeon server for lightweight preprocessing.  
    \item \textbf{Latency:} Per-sample inference takes 120--270 seconds for VideoLLM reasoning and 6--12 seconds for audio generation, depending on instruction complexity and number of events.  
\end{itemize}

\subsection{Limitations}
\label{appendix:limitations}

While EchoVidia improves controllability and semantic alignment, several limitations remain:

\begin{itemize}
    \item \textbf{Latency.} The multi-stage reasoning and synthesis pipeline increases inference time, particularly when many events are present.
    \item \textbf{Dependency on VideoLLM Awareness.} Event accuracy is bounded by the VideoLLM’s ability to perceive subtle motions or occluded interactions.
    \item \textbf{Diffusion Model Biases.} Stable Audio Open may generate artifacts or stylistic biases, especially for rare or highly specific sound textures.
\end{itemize}

Future work may integrate EchoVidia into end-to-end trainable architectures and extend the symbolic representation to capture richer auditory phenomena such as reverberation, multi-source interference, and dynamic spatialization.

%% file: supp/C-case_study.tex
\section{Motivation Illustration and Qualitative Examples}

Please see the video in project website. We present Figure~\ref{fig:teaser} with video illustration, and compare the performance of \ModelName{} with MMAudio-L-44.1kHz, ThinkSound, and HuanyuanVideo-Foley-XXL.